\documentclass[10pt,twocolumn,letterpaper]{article}

\usepackage{cvpr}
\usepackage{times}
\usepackage{epsfig}
\usepackage{graphicx}
\usepackage{amsmath}
\usepackage{amssymb}
\usepackage{subcaption}
\usepackage{array}
\usepackage{mathtools}
\usepackage{float}
\usepackage{color}
\usepackage{multirow}
\usepackage{algorithm}
\usepackage{algpseudocode}
\usepackage{bbm}

\def\cvprPaperID{***} % *** Enter the CVPR Paper ID here

\definecolor{dkmag}{rgb}{1.0,0.0,1.0}

\newcommand{\yasu}[1]{}
\newcommand{\hang}[1]{}
\newcommand{\yebin}[1]{}

\definecolor{gold}{rgb}{0.85,.66,0}

\newcommand{\mysubsubsubsection}[1]{\vspace{0.1cm} \noindent {\bf #1}:}

\usepackage[pagebackref=true,breaklinks=true,letterpaper=true,colorlinks,bookmarks=false]{hyperref}

\cvprfinalcopy % *** Uncomment this line for the final submission

\ifcvprfinal\fi
\begin{document}

%%%%%%%%% TITLE
\title{Turning an Urban Scene Video into a Cinemagraph}

\author{Hang Yan\\
  Washington University\\
  St. Louis, USA\\
  {\tt\small yanhang@wustl.edu}
  \and
  Yebin Liu\\
  Tsinghua University\\
  Beijing, China\\
  {\tt\small liuyebin@mail.tsinghua.edu.cn}
  \and
  Yasutaka Furukawa\\
  Washington University\\
  St. Louis, USA\\
  {\tt\small furukawa@wustl.edu}\\
}

  \maketitle
  %\thispagestyle{empty}

  %%%%%%%%% ABSTRACT
  \begin{abstract}
    This paper proposes an algorithm that turns a regular video capturing
    urban scenes into a high-quality endless animation, known as a
    Cinemagraph. The creation of a Cinemagraph usually requires a static
    camera in a carefully configured scene. The task becomes challenging
    for a regular video with a moving camera and objects. Our approach
    first warps an input video into the viewpoint of a reference
    camera. Based on the warped video, we propose effective temporal
    analysis algorithms to detect regions with static geometry and dynamic
    appearance, where geometric modeling is reliable and visually
    attractive animations can be created. Lastly, the algorithm applies a
    sequence of video processing techniques to produce a Cinemagraph
    movie. We have tested the proposed approach on numerous challenging
    real scenes. To our knowledge, this work is the first to automatically
    generate Cinemagraph animations from regular movies in the wild.
    % This paper proposes an algorithm to model dynamic appearances of static
    % indoor scenes from an RGBD stream. We focus on displays such as TVs,
    % monitors, and computer screens, which are the major sources of dynamic
    % appearances with static geometries in indoor environments.  A display is
    % always rectangular in 3D space, and our approach devises effective means
    % to exploit this prior throughout the process.  After generating quad
    % hypotheses in each frame via line segment detection and vanishing point
    % analysis, we iteratively use dynamic programming to extract all the
    % displays visible in a scene.  Our dynamic appearance model allows novel
    % ``time-dependent texture-mapping'', as an extension of standard
    % view-dependent texture-mapping, where a scene texture depends on time in
    % addition to the view-point. We tested our algorithm on various
    % real-world indoor environments, and compared our results against
    % competing baseline methods.
  \end{abstract}

  %%%%%%%%% BODY TEXT

  \section{Introduction}
%\IEEEPARstart{M}{arkerless} motion capture records 3D motion of an
%actor without marker or sensor suits.  It has a wide range of
%applications including human computer interaction, surveillance, gait
%analysis, and visual special effects.  Most of previous methods are
%limited in indoor studios, because of the required depth cameras
%(e.g. \cite{Shotton2011,wei2012accurate}) or tens of carefully
%calibrated cameras and chromatic background
%(e.g. \cite{Gall2009,Wu2013}).

Our world is dynamic. Imagine you are standing in the middle of Times
Square surrounded by constant noise, cars passing by, or flashy
billboards showing advertisements at every second. A fundamental
challenge in Computer Vision is to model and visualize dynamic
environments. Cinemagraphs, still photographs containing minor and
repeated animations~\cite{cinemagraph}, are one of the most successful
examples in capturing such scene dynamics. Their subtle animations are
effective in capturing the ``moment'' with striking visual effects.

The generation of high-quality Cinemagraphs have so far required static
cameras with carefully configured
scenes~\cite{cinemagraph,deanimating} or interactive
tools~\cite{cliplets}. No compelling techniques exist in the
automatic conversion of regular videos into Cinemagraphs.
%
% A milestone in Computer Vision research is automatic generation of
% Cinemagraph-style animations from regular movies in the wild.
%
Online photo storage services, such as Google Photos, automatically
produce short animations from user images and movies. However, their
animations are either a simple image slide-show or a trimmed movie
segment looping forward and backward unnaturally. 
%
% Our world is dynamic. Imagine you are standing in the middle of Times
% Square in New York City surrounded by cars passing by, people running around,
% and flashy billboards showing advertisements at every second. A
% fundamental challenge in Computer Vision is to model and visualize such
% dynamic environments,
% % \yasu{removed this, as this only refers to the rendering portion.}
% % which can be called "Tcime-dependent rendering",
% with applications ranging from art, digital mapping, to
% virtual tourism.
% %
% % \yasu{This is exactly the above research challenge (not additional
% % research problem).}
% % Additionally, the reconstruction of the dynamic scene also enables new
% % research problems such as dynamic content detection and retrieval.

We seek to make the first step towards automated Cinemagraph generation
from regular movies with moving cameras in the wild. The key insight
is that even subtle animations yield striking visual effects, where
our approach is to selectively and precisely segment regions that lead
to high-quality animations. In particular, we focus on urban
environments or night-time settings, where neon-signs, displays, or
flashy billboards decorate a scenery. Such a geometry is static,
making the modeling task significantly easier, while their appearance
adds effective dynamics to the scene visualization.
% , even being a fraction of a scene.
%
% Second, a static geometry can add effective dynamics through appearance
% changes, where the modeling task becomes significantly easiser. For
% instance, neon-signs, displays, or billboards are symbolic features for
% many urban environments or night-time sceneries.
%
% certain scene dynamics come from pure appearance variations with
% static geometry, such as neon-signs, displays or billboards, which are
% important elements in describing urban environments or night-time
% sceneries.
%
% Our inspiration comes from two award-winning papers in Computer
% Vision: {\it Dynamic Fusion} by Newcombe, Fox, and
% Seitz~\cite{newcombe2015dynamicfusion} that reconstructs non-rigidly
% deforming objects through RGBD streams, and {\it Scene Chronology} by
% Matzen and Snavely~\cite{matzen2014scene} that models the dynamic
% appearance of building facades over a long period of time through
% Internet Photo Collections. The former takes an arbitrary RGBD stream
% to model a dynamic geometry, but with little focus on appearance. The
% latter models dynamic appearances on static geometries, but is limited
% to a few landmarks where Internet photographs are available. This
% paper seeks to model dynamic appearances of static scenes for everyday
% indoor environments by using RGBD streams.

Formally, this paper turns a regular video capturing urban scenes into
a Cinemagraph-style animation in three steps.
%% Formally, this paper turns a regular video, containing a region with
%% static geometry and dynamic appearance, into a Cinemagraph-style
%% animation in three steps.
%
% A resultant movie adds a new dimension to the field of scene modeling
% and visualization, where animations, even being a small fraction of a
% screen, conveys vivid scene dynamics with striking effects.
%
% This paper proposes a system that turns a regular video into Cinemagraph
% renderings at one or more selected reference frames. This is a very
% challenging task, as Cinemagraphs are usually generated from static
% cameras with carefully configured scenes. Two ideas are at the heart of
% our approach.
%
% Second, we utilize state-of-the-art 3D reconstruction techniques to warp
% an input video into the viewpoint of a reference frame. The warped
% video allows effective spatio-temporal analysis of a scene to detect
% regions to be animated.
%
%The system then makes a cinemagraph rendering by only animating these
%regions while fixing other parts. (Section~\ref{render}).
%
%Our approach consists of three steps.
First, we utilize existing 3D reconstruction techniques to warp an
input video into the viewpoint of a reference frame. Second, novel
temporal analysis algorithms are applied to the warped video to give
regions where high-quality animations can be produced. These regions
have static geometry with repetitively or randomly changing
appearance. Third, we perform a sequence of video processing
techniques to generate high-quality animations for the segmented
regions, while fixing the rest of the pixels to the reference frame.

The contributions of this paper are two fold. The technical
contribution lies in the effective temporal analysis of noisy warped
videos to enable the segmentation and classification of visually
interesting regions. The system contribution is the fact that this is
the first effective system automatically generating Cinemagraph
animations from regular movies.

  \section{Related work}

\begin{figure*}[tb]
  \centering \includegraphics[width=\textwidth]{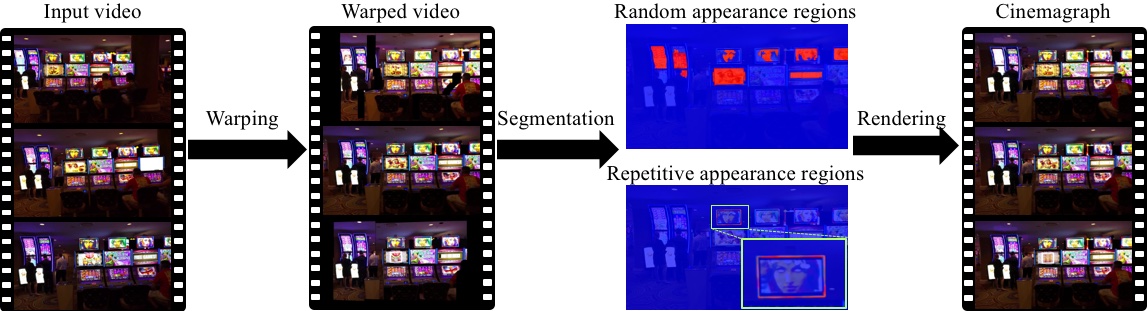}
  \caption{System overview. Given an input video clip (left), the
    system first warps all frames into the reference view (left
    middle) by computing camera poses and the reference
    depth-map. Regions that are visually attractive are selected,
    namely regions with static geometry and dynamic (random or
    repetitive) appearance (right middle). The system then creates a
    cinemagraph rendering by only animating those regions while fixing
    other pixels to the reference frame
    (right).}  \label{fig:overview}
\end{figure*}

Dynamic scene reconstruction has been a fundamental problem for Computer
Vision. Significant progress has been made for lab-environments, where
multiple calibrated and synchronized video cameras are the input.  A
successful system has been demonstrated for a human
body~\cite{starck2007surface}, a human face~\cite{DisneyFace}, or multiple
people with interactions~\cite{HanDynamic2014,collet2015high}.
% time-varying geometry
% reconstruction~\cite{starck2007surface,taneja2011modeling}, non-rigid
% SfM~\cite{bregler2000recovering} and scene flow
% estimation~\cite{vedula1999three,HanDynamic2014}.  However, all of these
% works either requires an expensive and controlled indoor
% setup~\cite{starck2007surface,HanDynamic2014} for high
% quality modeling and texture rendering \cite{collet2015high}, or only
% using only convenient capture devices but resulting in reconstruction
% and rendering quality far from practical
% use~\cite{bregler2000recovering,taneja2011modeling,vedula1999three}.
%
%
% Detection, segmentation and reconstruction of general dynamic scenes
% with moving objects remains an unsolved problem.
%
% Accurate segmentation of moving objects is still a very challenging
% problem~\cite{wang2005interactive}, even with an RGBD
% video~\cite{hickson2014efficient} or multiple calibrated and
% synchronized videos~\cite{djelouah2013multi}.
%
% Multi-view camera system has been an effective mean to recover
% high-fidelity 3D structure and motion~\cite{joo2014map}. However, the
% data acquisition requires a lab set-up with multiple calibrated and
% synchronized cameras.
%
%
Dynamic scene reconstruction from YouTube videos
%more convenient capture setups (e.g., a single video camera)
has been recently proposed~\cite{ji20143d}.
%
% 3D reconstruction of dynamic textures using a convenient capture setup
% like a single view camera is recently proposed
They jointly reconstruct the static background and dynamic foreground
objects.
%optimization the multiview segmentation of dynamic scene foreground.
However, the adopted visual hull reconstruction leaves noticeable
protruding artifacts on their models.
%
% as shown in the paper on reconstruction and rendering of a
% screen with dynamic texture, the adopted visual hull reconstruction
% mechanism causes plane artifact protrusions on the display.
Dynamic Fusion reconstructs non-rigidly deforming objects from an RGBD
stream, but does not focus on dynamic
appearances~\cite{newcombe2015dynamicfusion}.

More macro scale dynamics (e.g., scene changes over months or years) can
be detected by analyzing a set of images acquired by standard
cameras~\cite{WangKM15}, stationary surveillance
cameras~\cite{sakurada2013detecting}, vehicle-mounted
cameras~\cite{ulusoy2014image}, or community photo sharing
websites~\cite{matzen2014scene,martin2015time}.
%
% In longer time scale (changes happen in minutes, hours, days or
% years), Ulusoy et al. \cite{ulusoy2014image} and Sakurada et
% al. \cite{sakurada2013detecting} detected the change of the 3D scene
% and update the 3D reconstruction to form the history of a 3d scene
% over time from images.
%
In particular, the time lapse reconstruction~\cite{martin2015time} has
produced impressive spatio temporal 4d models. They are one of the
most successful examples of varying geometry and appearance over space
and time. However, they require massive amount of photographs,
limiting their applications to a small number of landmarks in the
world.  In contrast, we seek to realize Cinemagraph-quality dynamic
scene visualization from a single regular movie.
% acquired by a single camera.

The Cinemagraph creation has also been studied. Impressive results
have been obtained by semi-automatic
systems~\cite{cliplets,deanimating}, or with templated input
videos~\cite{portraits}. Automatic cinemagraphs
systems~\cite{cinemagraph11, cinemagraph12} create the mask for
animated regions by motion analysis. However, these systems are
successful only when the camera does not move and the scene is mostly
static.
A data-driven single-image approach has produced impressive
animations~\cite{kholgade20143d}, but the dependence on the database
currently limits their application ranges.
%
%All previous systems require carefully captured static
%videos.
In contrast, this paper proposes an automated approach for regular
movies with moving cameras.

  \section{System overview}

\begin{figure*}[!ht]
  \centering \includegraphics[width=\textwidth]{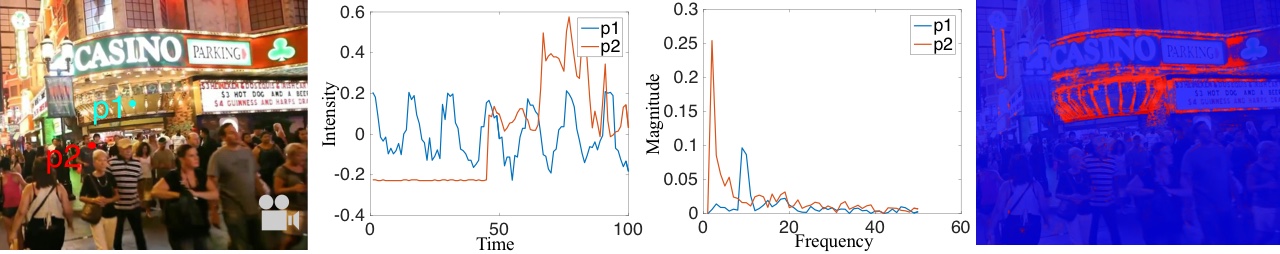}
  \caption{ Fourier analysis for repetitive pixels. Left: Two pixels
  $p1$ and $p2$ are marked for illustration in the reference frame. Left
  middle: the intensity patterns of two pixels over time. The
  intensities are preprocessed to have zero means. Right middle: the
  frequency magnitudes of temporal pattern. Since $p2$ has repetitive
  appearance, the peak occurs at a high frequency, while for $p1$ the
  peak occurs at a low frequency. Right: the resulting repetitive score,
  encoded in the red color channel.
    %% \yasu{Put axis
    %%   label such as ``Frames'' or ``Fourier coefficient?''. Currently, it
    %%   requires efforts to know what the middle 2 graphs are. There is enough
    %%   space to put some texts. Cannot think of a good terminology now
    %%   though. Maybe, frames and ``frequency index''???}
    %% \yasu{It is nice to show some comparisons later. For
    %%   example, what happens if we simply compute the average instead of
    %%   taking maximum in this formula. One example, showing ours and
    %%   this baseline, and the resultant segmentations should be good
    %%   enough.} \yasu{Use also transparency for the right result too?}
  } \label{fig:fourier}
\end{figure*}

This paper proposes a novel system that allows us to convert standard
movies with moving cameras capturing urban scenes into Cinemagraph
animations. Our system consists of three steps: warping, segmentation,
and rendering (See Figure~\ref{fig:overview}).  First, we use
Structure from Motion (SfM) and Multi-View Stereo (MVS) algorithms to
warp the input video into the viewpoint of a reference frame via a
depth-map based image morphing. Second, effective temporal analysis
and segmentation algorithms are applied on the warped video to give
regions that lead to good animations. Finally, we render a
high-quality Cinemagraph movie by a sequence of video processing
techniques in each segmented region to mitigate artifacts from
warping, while fixing the rest of the pixels to the reference frame.
The warping step is an application of standard techniques, while the
last two steps, in particular the segmentation step, exhibit technical
contributions in this paper.

  \section{Spatial alignment by image warping} \label{sec:warp}

% We use Structure from Motion (SfM) and Multi-View Stereo (MVS)
% algorithms to estimate camera poses and depthmaps. We expect a static
% part of the scene to make SfM work and MVS interpolates rough geometry
% over dynamic regions. We then warp the input video into the viewpoint of
% a reference frame via a depthmap-based image morphing. Reference frames
% are either manually selected or uniformly distributed across the video.

The video warping is based on standard 3D reconstruction techniques
with minor modifications for being robust against scene dynamics.
First, we use a SfM software TheiaSFM~\cite{theia-manual} to estimate
camera poses.  Given a reference frame $I_r$, we estimate a depth-map
based on 100 neighboring frames (i.e., 50 frames before and after) via
a standard MRF formulation. The range of scene depth at $I_r$ is
estimated from visible 3D points provided by SfM, with 1\% nearest and
farest points discarded for robustness. The inverse depth space is
then uniformly discretized into 128 labels.
%% The minimum and the maximum depths of SfM points visible from $I_r$
%% are used to uniformly sample 128 disparities, while discarding the
%% 1\% nearest and farthest points.

Following the idea in \cite{kang2004extracting}, we compute a matching
score between $I_r$ and each neighboring image, then sum up the best
half of these scores as the overall unary term to compensate for
occlusions and scene dynamics. The matching score is defined as one
minus the Normalized Cross Correlation over $5\times 5$ image patches,
truncated at $0.3$ for robustness.  The pairwise term is a truncated
linear function of the absolute label differences with a truncation at
4. We multiply 0.15 to the pairwise term. Given the estimated
depth-map, we warp all the neighboring frames into the viewoint of
$I_r$ via standard backward warping, while taking into account
occlusions via Z-buffering. Occluded pixels are ignored in the next
segmentation process and will be in-painted in the last rendering
process.

\section{Dynamic appearance segmentation} \label{sec:segmentation}
Carefully choosing regions to animate is the key to successful
Cinemagraph creation.
%
% Mixture of extensive scene dynamics and errors from the warping stage
% pose challenges to standard motion-based algorithms.
%
%Vasto amount of dynamics and errors from the warping stage pose
%challenges to standard motion-based algorithms.
%
% The key idea is that although the warping procedure implicitly treats
% all pixels as static, the temporal characteristics produced by dynamic
% geometry (e.g. pedestrians, vehicles) and static geometry but dynamic
% appearance (e.g. billboards, flashing signs) in the warped video are
% difference.
Our approach is to conduct temporal analysis on the spatially aligned
warped-video, while focusing on two types of appearances common in urban
scenes.

% In particular, we handle two types appearances
% common in urban scenes: 1) regions with randomly changing appearances
% and 2) regions with repetitively changing appearances.

\subsection{Randomly changing appearance} \label{sec:display}

Digital displays or billboards are popular visual attractions in urban
downtowns, night clubs, or store-fronts in shopping malls. Detection and
segmentation of displays pose challenges to existing techniques as they
could show arbitrary contents.
Our system detects these regions by 1) segmenting the warped video into
2D segments by a novel feature vector encoding characteristic appearance
changes, and 2) classifying these regions by a random forest trained
from manually annotated video clips.
%
%Typical examples of static geometry with randomly changing appearance
%in urban scene videos include digital displays and billboards, where
%%the randomness poses challenges to existing techniques. However,
%pixels colors inside the same dynamic region tend to change in the
%similar way.
% Our system detects these regions by 1) segmenting the
% warped video into 2D regions with novel pixel features that capture
% both spatial and temporal characteristics, and 2) classifying these
% regions by a random forest trained from manually annotated video
% clips.

\begin{figure*}
  \centering
  \includegraphics[width=\textwidth]{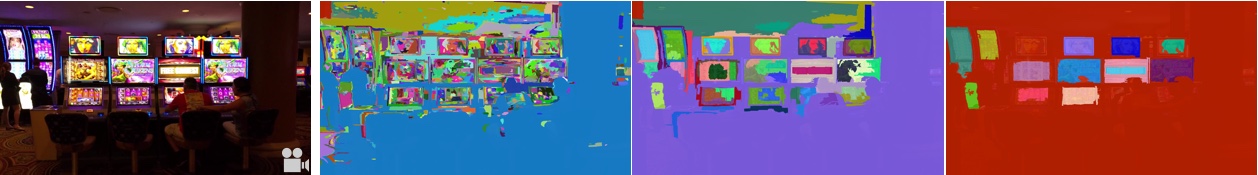}
  \caption{Hierarchical 2D video segmentation.  The left shows the
    reference frame. The right shows the segmentation results at
    0\%(lowest), 60\%, and 80\% hierarchy levels,
    respectively.}  \label{fig:my_seg}
\end{figure*}

\vspace{0.1cm} \mysubsubsubsection{Warped video segmentation} We
perform hierarchical bottom-up 2D segmentation of a warped video (See
Fig.~\ref{fig:my_seg}).~\footnote{The approach starts from pixels and
greedily merges the closest pair if the distance of their feature
vectors is below a threshold. It iterates between merging and increasing
the threshold until everything merges to a single segment. The original
algorithm~\cite{grundmann2010efficient} operates on 3D pixel volumes,
while our system performs 2D segmentation by treating each pixel as a 1D
array.} Our contribution lies in the novel distance metric for a pair of
spatial regions, defined as $D_{T} + 0.1 D_{A}$. The metric utilizes the
warped video to analyze temporal appearance changes $D_{T}$ in addition
to the absolute appearances $D_{A}$.

\begin{figure}[!tb]
  \centering \includegraphics[width=0.49\textwidth]{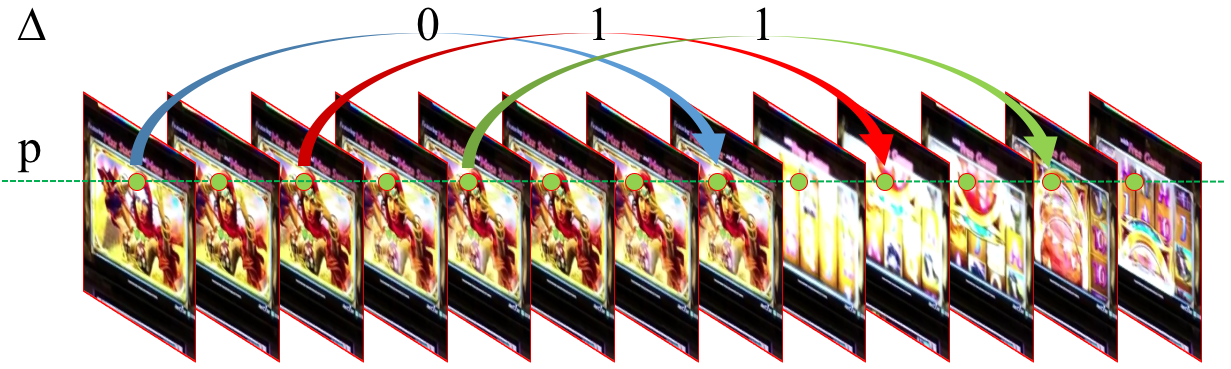}
  \caption{ Our temporal binary pattern encodes the temporal
    characteristic of a pixel (p) by checking once in every $\alpha$
    frames if there will be significant color change after $\beta$
    frames. The figure illustrates a case when $\alpha = 2$ and $\beta
    = 7$.}  \label{fig:transition}
\end{figure}

The inspiration of $D_{T}$ comes from locally binary
pattern~\cite{calonder2012brief,ojala1994performance} for feature
matching, which we employ in the temporal domain. Let $I_p(f)$ be the
mean color of a pixel $p$ (or a region after merging) at frame $f$ in
the warped video, where we will drop $p$ from the notation below for
simplicity. Our binary temporal pattern descriptor checks once in
every $\alpha$ frames if there will be significant color change
$\beta$ frames ahead (See Figure~\ref{fig:transition}). More
precisely, the $i_{th}$ bit of the binary pattern is defined as
\begin{align} \label{eq:descriptor}
  \Delta(i) = \mathbbm{1}(\lVert I(\alpha i + \beta) -
  I(\alpha i) \rVert_2 > \theta).
\end{align}
$\mathbbm{1}$ denotes the indicator function and checks if the color
difference is more than $\theta = 100$.

To capture both local and long term appearance patterns, we use two
sets of parameters: $\alpha_1=4,\beta_1=4$ for $\Delta_1(i)$, and
$\alpha_2=2,\beta_2=N/2$ for $\Delta_2(i)$.  The final binary
descriptor is the concatenation of the two:
\begin{eqnarray*}
 T=[\Delta_1(0), \Delta_1(1),\ldots, \Delta_2(0),\Delta_2(1)\ldots].
\end{eqnarray*}
 The distance between two binary temporal descriptors $D_{T}$ is
 computed by the Hamming distance of the two binary vectors normalized
 by the feature dimension.

$D_A$ measures the appearance difference between two segments by
computing the one minus the normalized cross correlation of two
histograms. The histogram is constructed from the LAB values of all
pixels inside the 2D region from all frames, with 8 bins per channel.

We start the process with the initial merging threshold set to 0.2 and
increase it by a factor of 1.5 every time. We use segmentation results
at three different levels of granularity to generate (overlapping) image
segments for robustness, in particular, at 60\%, 70\% and 80\%
levels~\cite{grundmann2010efficient}.

\vspace{0.1cm} \mysubsubsubsection{Classification} We build a binary
random forest classifier based on appearance, temporal changes, shape,
and position features.  We have obtained the training data by
downloading stationary video clips of popular urban scenes from
YouTube. Then, we have manually annotated 2D regions such as displays
and billboards.
% with dynamic appearance but static geometry such as displays and
% billboards.
Please refer to the supplementary material for detailed feature design
and training process. At test time, we pass all segments from three
granularity levels into the classifier and take the union of all
positive segments. We ignore mostly invisible segments, that is, if more
than half the pixels are invisible (i.e., project outside the view or fail in
the Z-buffering test during warping) in more than half the neighboring frames.
% Small and short visibility holes are in-painted in the rendering stage.

\subsection{Repeatedly changing appearances} \label{sec:flashing}

Repeated advertisements or flashing neon-signs are also symbolic
structures in many urban scenes, especially at night time. Due to the
fact that these regions are often small and isolated, standard motion
analysis and segmentation algorithms perform poorly. We propose a simple
but powerful temporal analysis algorithm based on Discrete Fourier
Transform (DFT) to recognize these pixels.

For each pixel of the warped video, we compute the 1D DFT of its
intensities over all frames, then conduct a frequency analysis (See
Fig.~\ref{fig:fourier}). For ideal cyclic intensity patterns, we should
observe a clean peak among high frequency components, while low
magnitudes at low frequencies. To be robust against errors from warping,
instead of computing a single score using all the $N$ frames, we look
for the optimal interval of at least $N/2$ frames. More precisely, we
compute the repetitive-ness score of a pixel from frame $i$ to $j$ as
\begin{equation*}
  C_{rep}(i, j) = \frac{\max_{k> \tau}{|F_k|}}{\max_{k\le \tau}{|F_k|}}.
\end{equation*}
$|F_i|$ denotes the magnitude of a DFT component.~\footnote{We only use
the first half of the DFT coefficients as their magnitudes are symmetric
for real-valued arrays. We also discard the direct component by
subtracting the mean intensity before DFT.} $\tau$ is the boundary
between the low and high frequency components, which is set to 4
throughout the experiments. The final score is the maximum over all the
possible frame intervals containing at least $N/2$ frames:
\begin{equation}
  C_{rep} = \max_{(j - i) \ge N/2} C_{rep}(i, j). \label{eq:flashy}
\end{equation}
We animate a pixel if 1) its score (\ref{eq:flashy}) is greater than
2.5 and 2) its 80th percentile of intensities over all frames is
greater than 127.
%% We discard pixels whose scores (\ref{eq:flashy}) are less than
%% 2.5. Finally, we also filter out dark pixels, whose 80th percentile
%% intensity in the warped video is less than 127, and have the remaining
%% pixels animated in the Cinemagraph movie.

  \section{Cinemagraph rendering}

% \mysubsubsection{Animation:} We generate a Cinemagraph movie by
% rendering an animation in each segmented region. We perform hole-filling
% to inpaint occluded pixels and stabilization to compensate for errors in
% camera poses or scene geometry.  Lastly, a low-rank matrix approximation
% technique suppresses rendering artifacts. A generated Cinemagraph
% provides an endless animation with vivid dynamics of a scene.

Our system renders Cinemagraph animations in the detected regions
using frames from the warped video, while keeping the remaining pixels
fixed to the reference frame. The repeated pattern often consists of
small segments and the animation in the optimal interval computed by
the formula~(\ref{eq:flashy}) looks natural without any
post-processing.

Random appearance segments are more challenging. We first in-paint
visibility holes by Laplacian smoothing over space and time. Next, we
apply geometric stabilization by homograph warping. More concretely,
feature tracks are generated~\cite{goodfeatures,opticalflow} and
filtered by the constraints that 1) the track has to last for at least
10 frames and 2) the standard deviation of the tracked pixel coordinates
must be less than 2 pixels in both x and y.  Linear least square are
used to compute the homography warping from each frame to the reference.

As in existing
literature~\cite{bennett2007computational,martin2015time}, we apply
intensity regularization. This is crucial for our warped video, which
suffers from severe rendering artifacts. Standard techniques such as
temporal median filtering~\cite{bennett2007computational} or global
least squares optimization~\cite{martin2015time} have produced
compelling results for many of our examples. However, they show two
typical failure modes when a segment exhibits rapid optical flow motions
and/or abrupt temporal changes. First, they over-regularize high
frequency temporal signals. Second, inconsistencies arise across pixels
due to the lack of spatial regularization.

Our approach is to represent pixel colors of a segment throughout the
frames as a 2D matrix and obtain a low-rank approximation.  This method
achieves moderate temporal and spatial regularization.
RPCA~\cite{wright2009robust} is the choice of our machinery, which has
been successfully used for various image and video analysis
tasks,
% such as background extraction,
but not for high-quality movie rendering to our knowledge.

\begin{figure}
  \centering
  \includegraphics[width=0.49\textwidth]{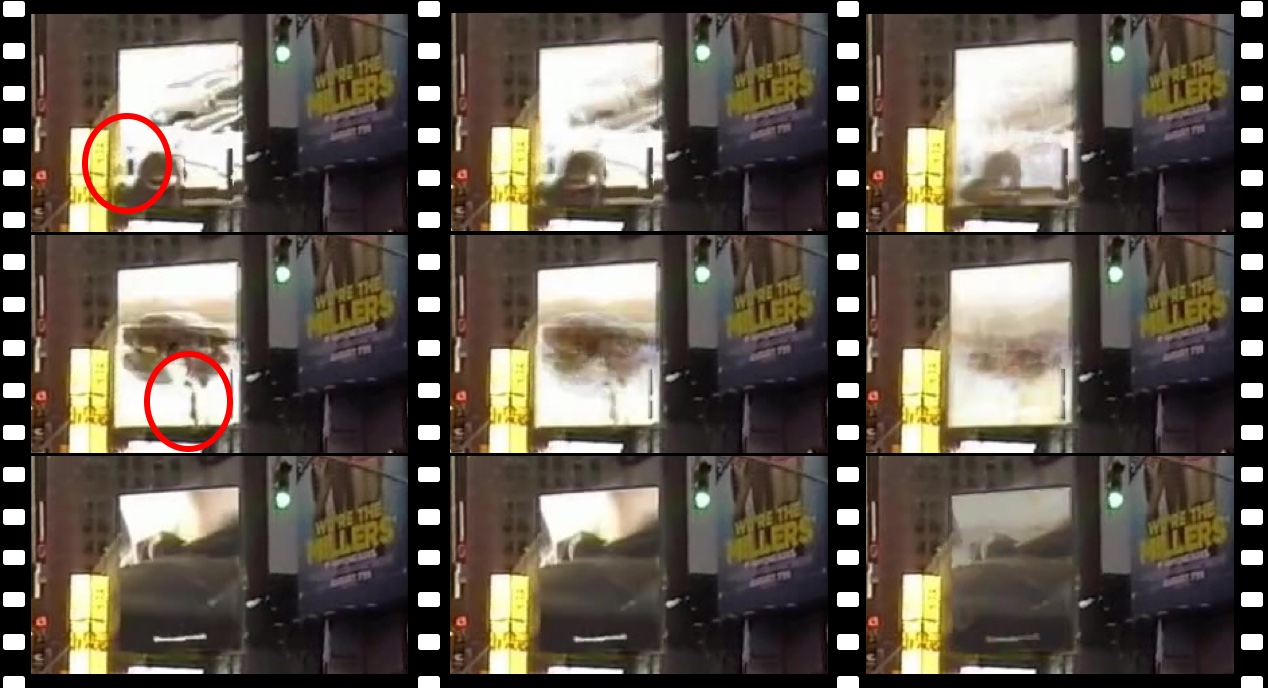}
  \caption{Adaptive appearance regularization. Left: with
    $\lambda=0.02$. Notice the occluders inside the red
    circles. Middle: output of RPCA with $\lambda_r=0.011$, which is
    automatically selected. The occluding artifact is
    mitigated. Right: output of RPCA with $\lambda_r=0.005$. The
    frames are overly smoothed.}
  \label{fig:adaptive}
\end{figure}

More concretely, we concatenate pixels of a segment in a single frame
to a row vector, and stack them across the frames to form a matrix
$P$. We use Accelerated Proximal Gradient~\cite{lin2009fast} method to
minimize the standard RPCA formulation:
\begin{eqnarray}
\lVert A \rVert_* + \lambda \lVert E \rVert_1\quad \mbox{subject to} \quad
A+E=P.
\end{eqnarray}
%% \begin{align}
%%  \min_{A,E}{\lVert A \rVert_* + \lambda\lVert E
%%     \rVert_1}  \\
%%  s.t.\;\; A + E = P \label{eq:RPCA}
%% \end{align}
The minimization of the nuclear norm $\lVert A \rVert_{*}$ achieves
spatial and temporal regularization. We solve the problem for each
channel independently and rearrange $A$ as the output pixel values.  We
have found that it is important to adaptively tune the scalar weight
$\lambda$ depending on the video content, which varies significantly
across examples.  Intuitively, a rich video content with fast optical
flows or temporal changes should still have large nuclear
norm. Therefore, we set $\lambda$ to be proportional to the
``richness'' of the video content,
% amount of drastic changes of the video inside the segment,
characterized by $\Delta_1$ from Section~\ref{sec:display}. More
precisely, we set $\lambda$ to be 0.005 + 0.015 $\gamma$, where $\gamma$
is the number of '1' in $\Delta_1$ divided by its dimension (See
Figure~\ref{fig:adaptive}).

Repetitive appearance regions are naturally cyclic, and we simply render
them repeatedly in an interval found by the optimization
(\ref{eq:flashy}). For random appearance regions, we create loops by
playing the video forward and backward.

\begin{figure}
  \centering
  \includegraphics[width=0.49\textwidth]{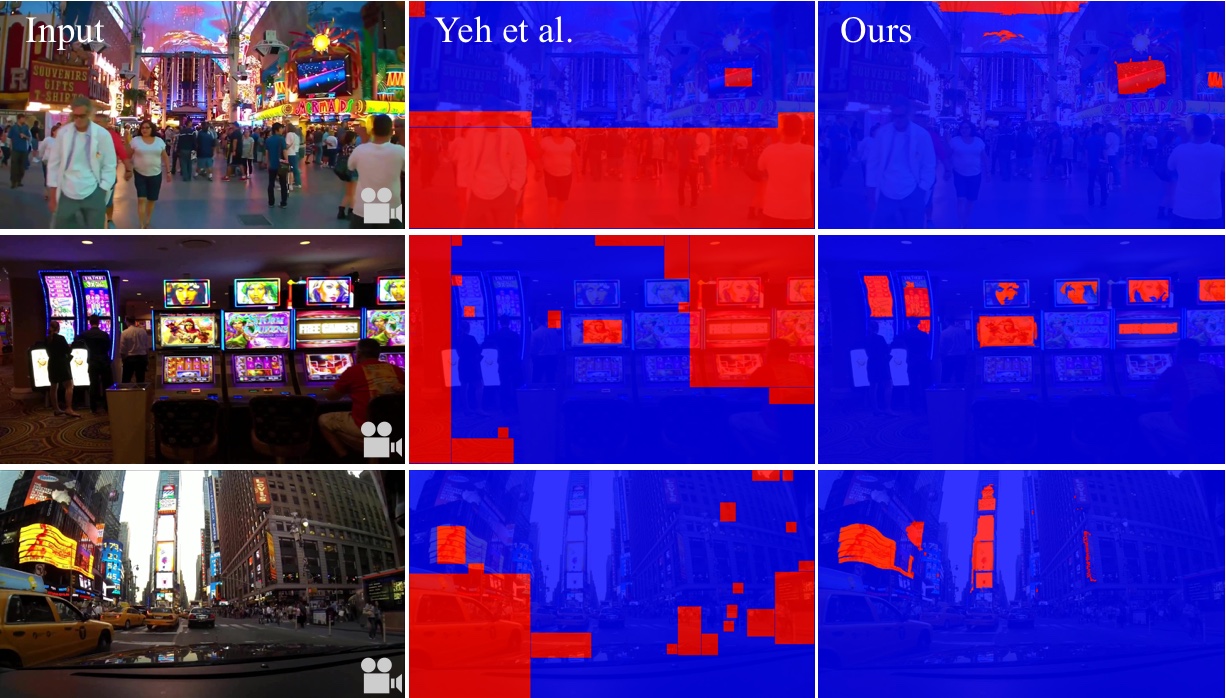}
  \caption{ Segmented regions for Cinemagraph animation by Yeh et
    al.~\cite{cinemagraph12} and our approach.  The method by Yeh et
    al. assumes a static camera as in any other Cinemagraph creation
    methods. It simply identifies a region with large optical flow
    motions, and fail to identify effective regions for Cinemagraph
    animations. }
 \label{fig:compare_cinemagraph}
\end{figure}

\begin{figure*}[tb]
  \centering
 \includegraphics[width=\textwidth]{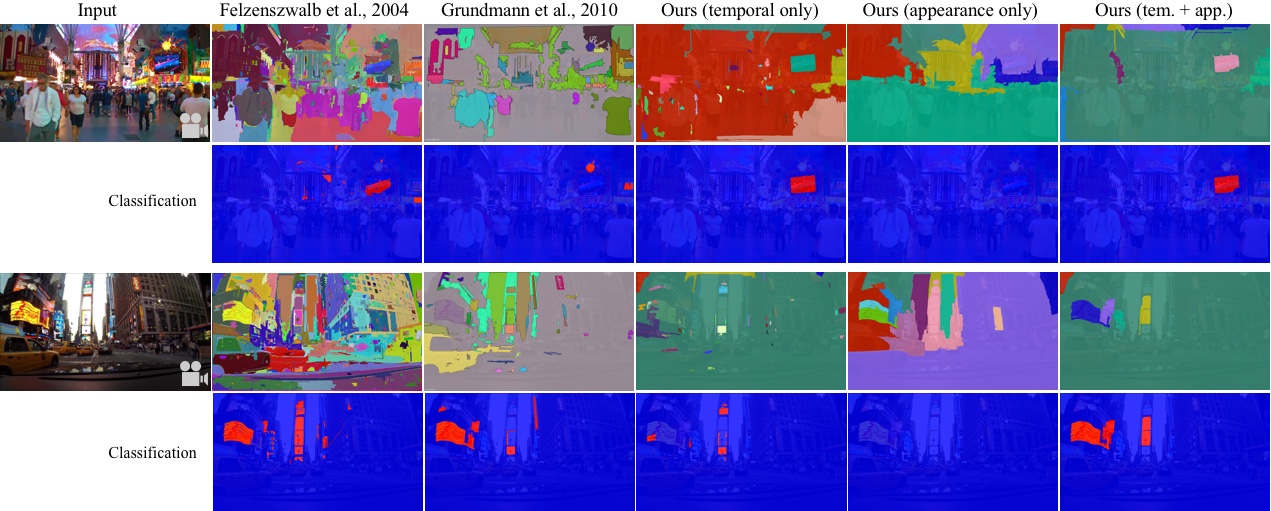}
 \caption{ We evaluate five different segmentation algorithms on two
 examples. For each example, the segmentation result is shown in the
 first row and the corresponding classification result is shown in the
 second row. The combination of the temporal and appearance (temp. +
 app.) information allows us to effectively segment regions that lead to
 good animation. Algorithms of Felzenszwalb et al.~\cite{SuperPixel}
 (with threshold parameter set to 500) and Grundmann et
 al.~\cite{grundmann2010efficient} lacks in temporal appearance
 information and fail to group pixels in the display in the top example.
 %, since they look spatially different.
 Only using the temporal information (temporal only) produces incomplete
 segments for partially dynamic displays (yellow display on the left of
 the bottom example).  Only using appearance information (appearance
 only) fails to capture display segment as in 
 Felzenszwalb et al. and Grundmann et al.
 Segmentation result from 80\% hierarchy are shown in the right four columns.
 %For the experiments in the right 4 columns, we show the result from
 %80\% hierarchy.
 }  \label{fig:compare_seg}
\end{figure*}

\begin{figure*}[tbh]
  \centering
  \includegraphics[width=\textwidth]{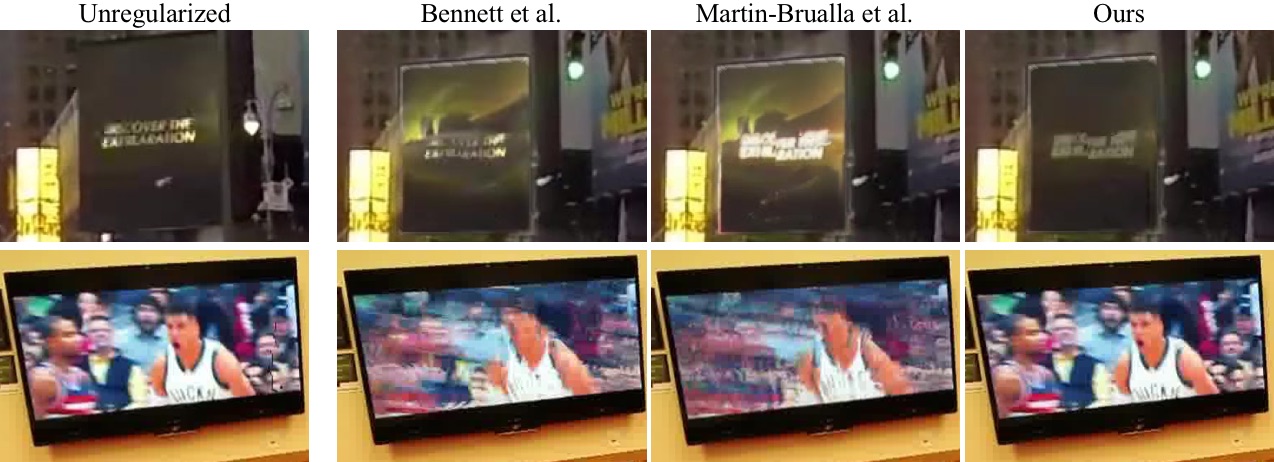}
  \caption{Comparison of different intensity regularization
  algorithms. Left: input frame. Right: rendering using temporal median
  filter with radius of 5, global least-square optimization with the smoothing
  weight 50, and our adaptive RPCA. The first two algorithms regularize
  each pixel independently, causing inconsistency across pixels under
  fast motion. In contrast, our algorithm jointly regularize all pixels
 inside a region.
 }
  \label{fig:compare_rendering}
\end{figure*}

\begin{figure*}[thb]
  \centering \includegraphics[width=\textwidth]{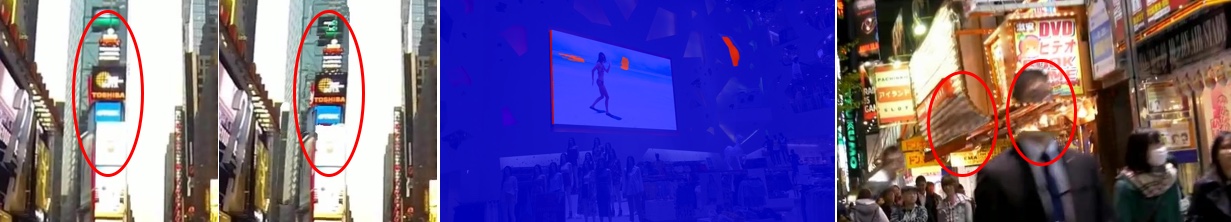}
  \caption{Failure cases. Left: the appearance inside the red oval is
 highly distorted in the second example
 % on the right half is distorted compared with the input frame on
 %    the left half. This distortion is caused 
 due to geometric errors from SfM and stereo. Middle: we fail to detect
 this display, which is mostly static with minimal appearance
 dynamics. We need more training data.
 % region is mis-classified, as this display is too static
 % . Due to the limited amount of training
 % videos, our random forest is not good at detecting display-like
 %    regions with only subtle changes.
 Right: defects caused by segmentation and large occluders. Although we
 handle small and fast occluders by intensity regularization, large and
 slow occluders still cause visual defects in the final rendering.}
 \label{fig:failure}
\end{figure*}

\section{Experimental results} \label{sec:experiments}
We have implemented the proposed system in C++ and used Intel Core I7
CPU with 32GB RAM and NVIDIA Titan X GPU (for stereo matching score
evaluation). We have downloaded various footages from YouTube such as
walk-throughs or drive-throughs of urban scenes. We have also recorded
walk-through videos by ourselves. Most of the input videos are 10
seconds long (i.e., 300 input frames), while some last for 2 minutes.
Our movie collections span indoors/outdoors, day/night, and various
places such as urban downtowns, city streets, casinos, shopping malls,
or university buildings.
%Note that while primarily designed for architectural scenes, our system
%are also applicable to other types of scenes that have static geometry
%and dynamic appearance, such as waterfalls or fountains.
Notice that SfM processes all the input frames, but our algorithm only
needs a reference frame and 100 neighboring frames.
The running time of our system after SfM step ranges from thirty minutes
to an hour, depending on the frame resolution and the number of display
segments.

\begin{figure*}[tb]
  \centering \includegraphics[width=\textwidth]{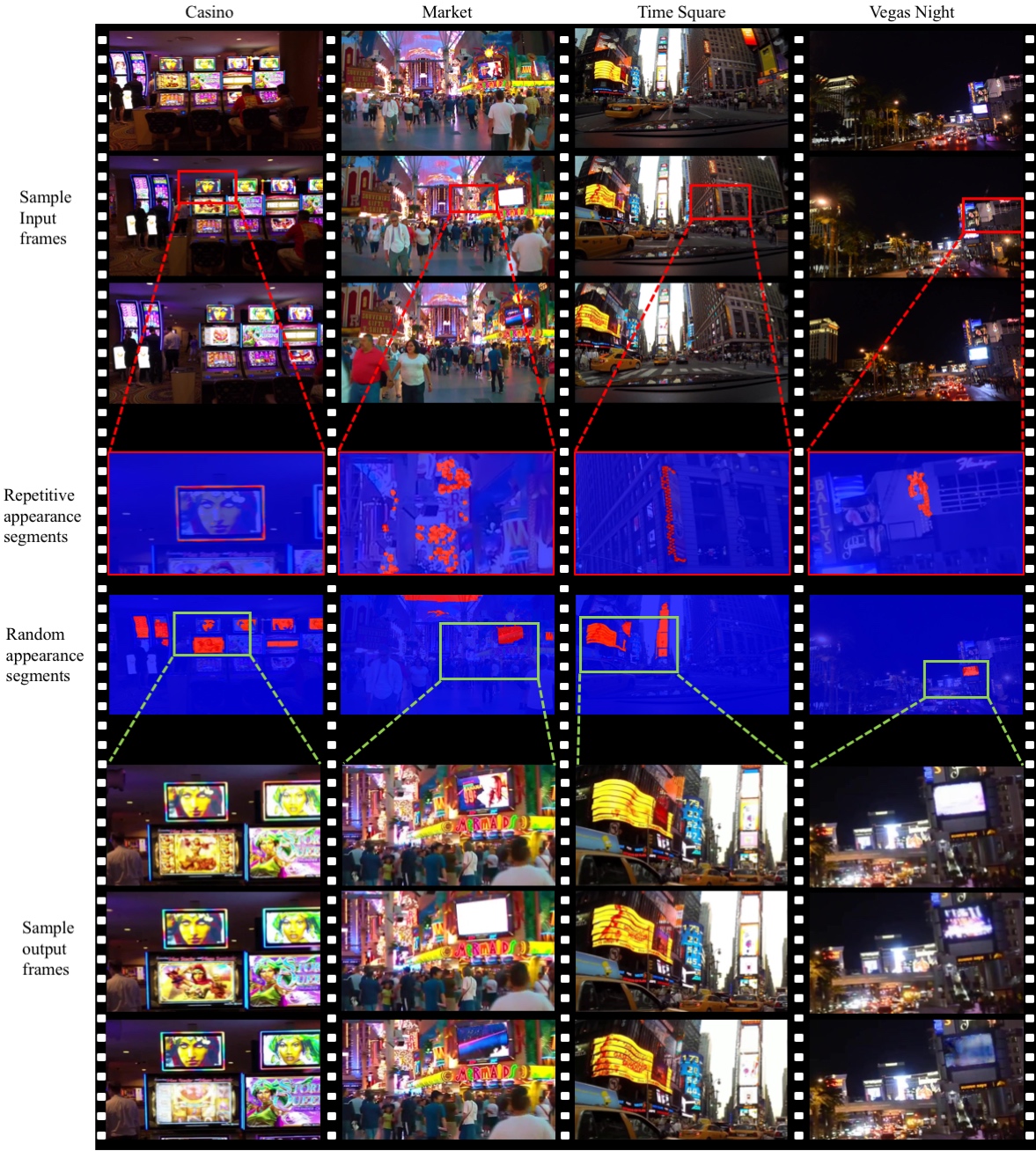}
  \caption{ A column shows sample input frames (the middle frame as
    the reference), a repetitive region, a randomly changing region,
    and sample output frames. Outputs for repetitive pixels and output
    frames are cropped to the red and green bounding boxes to better
    illustrate the details. Please see our supplementary video for the
    full assessment of our results and more examples. }  \label{fig:result_pipeline}
\end{figure*}

Figure~\ref{fig:result_pipeline} shows four of the input videos, segmentation
results of the two algorithms, and the output Cinemagraphs.
%, for four of our examples.
Our system enables automatic high-quality Cinemagraph creation from
videos with moving cameras in the wild, where all the existing
approaches require a static camera with a clean scene and/or human
manual interventions, to our knowledge. Please refer to the
supplementary material for complete experimental results and movies.

% put the large figure at last

%
% Our system is able to create visually attractive renderings from a
% large variety of inputs, including day time, night time, indoor
% and ourdoor. See Figure~\ref{fig:result_pipeline} for selected
% result. For more results and demo videos please refer to the
% supplementary material.
%

%We have compared our results against various other competing methods to
%assess the technical contributions of this paper.
%, namely the segmentation and the rendering algorithms via
%spatio-temporal analysis of warped videos.
Figure~\ref{fig:compare_cinemagraph} demonstrates that our segmentation
process effectively identifies regions that lead to high quality
animations.
Since no automated Cinemagraph generation method exists for a general
movie, we have supplied our warped videos to an existing algorithm
assuming a static camera~\cite{cinemagraph12} for comparison.
However, their algorithm simply looks at a rectangular regions with
large optical flow motions, and cannot handle severe rendering artifacts
or rich dynamics in our movies.

% moving objects, and produce highly errorneous segmentation results
% that lead to bad animations.
%The figure shows their result, while  This is done by repeatedly
%running the algorithm while excluding previously selected segments.
%
% Our warped video contains moving objects and severe
% rendering artifacts, and their method failed in identifying effection
% regions for animation.
%
% Existing automatic piplines mostly rely on motion analysis. They all
% assume that the input video is mostly static and will fail on our input
% videos. We compare our system with the system of Yeh et
% al.~\cite{cinemagraph12}. See
% Figure~\ref{fig:compare_cinemagraph}. Their system identifies animating
% segments as the regions with large flow magnitudes and can detect
% display regions with rough rectangles. However, their system is
% vulnerable to motions from pedestrains vehicles and fails to detect
%regions with repetitive appearance.
% We further improve the rendering inside display-like regions by
% temporal regularization. We adopt the similar idea
% from~\cite{timelapse}, which treats the RGB values of pixels over all
% frames as variables and achieves regularization by
% optimization. However, we use fundamentally different formulation.
%
%The regularization is apply to each display-like region
%independently. We first construct a matrix $P$ by using all pixel
%values

Figure~\ref{fig:compare_seg} shows the effectiveness of our new pixel
distance metric for display segmentation. The feature vector allows us
to extract image segments that have similar temporal changing
patterns. We compare our algorithm with Felzenszwalb et
al.~\cite{SuperPixel} on the reference image and Grundmann et
al.~\cite{grundmann2010efficient} on our warped video. Both methods
fail to extract segments that have similar temporal appearance
characteristics.
% but difference spatial appearance.
% The figure also demonstraWe also evaluate the effectiveness of the two terms

% both $D_T$ and $D_A$ parts in the
% new distance metric by dropping either of them.

% We first compare our 2D video segmentation to the hierarchical video
% segmentation algorithm~\cite{grundmann2010efficient}. Since our
% algorithm operates on 2D domain, we also make comparison to the
% original graph-based image segmentation
% algorithm~\cite{SuperPixel}. In the latter case, we only use the
% reference image and ignore temporal information.  See
% Figure~\ref{fig:compare_seg} for results. Notice that our algorithm is
% able to group pixels inside the same display region together, even
% though they have different appearance.

Figure~\ref{fig:adaptive} shows the effectiveness of our adaptive
appearance regularization, where we control the low-rank
regularization weight depending on the richness of the
video content.
%dynamic appearances.
Figure~\ref{fig:compare_rendering} shows our rendering results against
two standard intensity regularization techniques: temporal median
filter~\cite{bennett2007computational} and global least-squares
optimization~\cite{martin2015time}. These methods have no spatial
regularization (i.e., per-pixel operation), causing inconsistencies
across pixels in the presence of fast optical flow motions. Our low-rank
approximation technique outperforms in such cases with moderate
spatio-temporal regularization.
%, and similar results in the other cases.

\vspace{0.1cm} \mysubsubsubsection{Applications} The capability to turn
general videos into Cinemagraph animations opens up potentials for
novel applications.
%, for instance, in the space of digital mapping and virtual
%advertisements.
For instance, in the field of scene visualization, image-based
rendering navigation has become the golden standard (e.g., Google Maps
Street View)~\cite{google-maps,matterport}, where a user looks at a
real photograph at one location, and jumps between locations via
transition rendering.
 % to navigate around a scene.
However, photographs are all static without any dynamics in these
systems.
While directly serving videos might be a solution to visualize scene
dynamics, they require a lot more data space/transfer and constraint
the navigation strictly on the video path.
Replacing images with Cinemagraphs allows one to experience scene
dynamics at each location as well as free navigation in a scene.
Cinemagraphs animate only a fraction of an image and requires minimal extra
data space.
In particular, we demonstrate this next-generation Cinemagraph-based
rendering navigation, by taking a long walk-through video, generating
Cinemagraph animations at sub-sampled frames, then form a navigation
graph by connecting these frames. Furthermore, it is also easy to 
%our segmentation allows us to easily
replace animating contents by another media for virtual advertisement,
% for example, experiment for
% the first step towards
%virtual advertisement,
which might prevail in the near future with the emerging VR and AR.
Please see the supplementary video for examples.

%for example, turning geo-tagged user movies into Cinemagraph animations
%at subsampled locations
%
%Street-View style immersive image-based scene navigation with dynamics.

\section{Limitations and future work} % \label{sec:failure}

This paper proposes the first effective system that turns regular movies
with moving cameras into high-quality Cinemagraph animations at urban
environments.
Automatic Cinemagraph creation from regular video is still a very
challenging problem, and we have observed a few major failure modes
(See Figure~\ref{fig:failure}). First, our system expects that SfM
utilizes a static part of a scene to produce camera poses and MVS
interpolates rough geometry over dynamic regions. These assumptions might
%However, SfM and/or MVS might
fail at highly dynamic regions, causing unnatural distortions in the
animated contents. The second failure mode is in the classification. Our
training data come from movie clips by stationary cameras, which look
different from the warped videos. We need more training data,
potentially, annotating the warped videos from our algorithm for
training.
% are limited the amount and look different from the warped videos.
%
The last failure mode is in the rendering.
% Lastly, the rendering step often faces challenges.
While RPCA is very powerful in suppressing artifacts, it still fails
under the presence of severe occluders, such as the long appearance of
pedestrians in front of a camera. Utilization of semantic segmentation
techniques is our future work to make our system further robust against
occluders.

This paper makes a first important step towards automated high-quality
dynamic scene visualization from regular movies by mass consumers. We
hope that this paper will fuel a round of new research, tackling more
diverse set of dynamics in our world.

%Two important future directions are ahead of us: Firstly, we can utilize
%results from multiple reference frames to improve the robustness of the
%system by simple voting or joint optimization. Secondly, the new
%temporal analysis methods are specialized for urban scenes.
%It would be interesting, though challenging, to expand the idea to more
%types of scenes.

%improving the robustness of the system and expanding the scene types.

%% Due to the inherent difficulties and ambiguities of the problem, we also
%% observe failure cases during experiments. See Figure~\ref{fig:failure}
%% for details.

%% In the case of multiple reference frames across the video, our system
%% does not enforce consistency between reference frames. However, Since
%% our system estimates depth maps for all reference frames, this
%% consistency can be easily enforced by voting or joint optimization. We
%% take this as a future improvement.

  % \section{Conclusions} \label{sec:conclusions}
% This paper proposes a fully automatic system for creating Cinemagraph
% rendering from input videos captured by normal moving camera at urban
% scenes. The system is powered by the noval temporal analysis and
% regularization techniques. Experiments have shown the effectiveness
% and robustness of the system.

  %% Acknowledgement
  \section*{Acknowledgement}
  This research is partially supported by National Science Foundation under grant IIS 1540012 and IIS 1618685, and Microsoft Azure Research Award.

  \clearpage
  %%%%%%%%% TITLE
\begin{table*}[t]
  \centering
  \begin{tabular}{|p{0.09\textwidth}|p{0.14\textwidth}|p{0.63\textwidth}|p{0.04\textwidth}|}
    \hline
    Category & Feature & description & Dim \\
    \hline
    \multirow{3}{*}{Appearance} & RGB mean & Mean RGB values. & 3\\ \cline{2-4}
    & RGB variance & Variances of RGB values.& 3\\ \cline{2-4}
    & LAB histogram & Histogram in LAB color space with 8 bins for each channel. & 24 \\
    \hline
    Gradient & BoW & Bag-of-words descriptors constructed by K-means clustering on HoG3D
    descriptor~\cite{klaser2008spatio} extracted from regular grid points. & 100\\
    \hline
    \multirow{5}{*}{Shape} & Area & Ratio of the 2D area against the area of the entire frame. & 1\\ \cline{2-4}
    & Convexity & Ratio of the 2D area against the area of its convex hull. & 1 \\ \cline{2-4}
    & Rectangleness & Ratio of the 2D area against the area of the minimum bounding box. & 1\\ \cline{2-4}
    & Aspect ratio & The aspect ratio of the minimum bounding box. & 1 \\ \cline{2-4}
    & Number of edges & The number of edges of an approximated 2D
    shape. The approximation error is set to 1\% of the smaller dimension of the frame. & 1\\
    \hline
    \multirow{2}{*}{Position} & Centroid & The position of the
    centroid of the segment. & 2\\ \cline{2-4}
    &Bounding box & Minimum/maximum x/y position, normalized by width
    and height. & 4\\
    \hline
  \end{tabular}
  \caption{The feature vector used for random forest.}
  \label{tab:fv}
\end{table*}

\section*{Supplementary material}
The supplementary document provides more details on the classification
algorithm for randomly changing appearances.

We use a binary random forest to identify segments of randomly changing
appearance that lead to high quality animations. 
% We build a display classifier by a binary random forest with
Table~\ref{tab:fv} gives the full specification of our feature vector
that encodes appearances, temporal, shape and position information in
an 2D segment.

We have obtained the training data as follows. First, we have downloaded
stationary video footages from YouTube and manually annotated 2D display
regions at pixel levels. Second, we have run the same video segmentation
algorithm and generate segments. Lastly, we label each segment as a
positive (resp. negative) sample if more (resp. less) than 80\% the
segment overlaps with the annotated display pixels.
We then train a random forest (100 decision trees with depth 10) by
using 48 video clips for training and 12 video clips for validation. The
training and validation accuracy are 98.0\% and 93.0\%,
respectively. After obtaining the optimal hyperparameters, we merge the
validation set into the training set and re-run the training.
%
% After segmenting the warped video into a hierarchy of 2D segments, we
% build a binary random forest to classify segments into display regions
% or non-display regions. We construct a feature vector for video
% segments from appearance, gradient and shape features. See
% Table~\ref{tab:fv}.
%

% We compose our training set from Internet video clips captured by
% fixed camera. For each video clips, we manually annotate display-like
% regions. We extract training samples (2D video segments) by running
% segmentation algorithm in the previous section. Segments from all
% hierarchy levels are used. A segments is assigned the positive label
% if over half of the pixels are inside display annotations. We use 48
% video clips for training and another 12 video clips for
% validation. The trained random forest consists of 100 decision trees
% with tree depth 8. The achieved training accuracy and validation
% accuracy are 98.0\% and 93.0\%, respectively.

At test time, we classify each segment by the trained random
forest. Since image segments are obtained from the three levels of
granularity, each pixel belongs to three different image segments. We
treat a pixel to be a dynamic appearance segment, if at least one of the
three segments is classified as a display. After finding the connected
display components, we discard too small (less than 50 pixels) or too
large (more than 30\% of the frame) segments. Finally, we also discard
segments that are mostly occluded in the other views and do not likely
produce good animations.  More precisely, we discard a segment if more
than half the pixels are occluded in more than half the neighboring
views based on the visibility test conducted in the video warping
process.

% During testing, we pass segments from 50\%, 65\% and 80\% hierarchy
% levels to the random forest. A pixel is considered to be inside
% display-like regions if at least one of the three segments it resides
% in is classified as positive. We then extract connected components
% from the pixel level binary result. Components with too small area
% (below 50 pixels) and too large area (over 30\% of total pixels) are
% discarded. We also discard segments with over half of the pixels
% occluded at over half of views.

% \section{Rendering}
% After finding the connected display components, we discard segments
% that are too small (less than 50 pixels), too large (more than 30\% of
% the frame), or have too many rendering holes (more than half the pixels
% in more than half the neighboring views).

  \clearpage
  %%%%%%%%% BODY TEXT
      {\small
        \bibliographystyle{ieee}
        \bibliography{main}
      }

\end{document}